\documentclass{article}

\usepackage{microtype}
\usepackage{graphicx}
\usepackage{subcaption}
\usepackage{booktabs} 

\usepackage{stix}
\usepackage{color,colortbl}
\definecolor{lightgray}{gray}{0.9}
\usepackage{array,multirow,multicol}
\usepackage{amsmath,stmaryrd,amsfonts,amssymb,wasysym}
\newcounter{magicrownumbers}
\newcommand\rownumber{\stepcounter{magicrownumbers}\arabic{magicrownumbers}}
\usepackage{hyperref}

\newcommand{\argmax}[1]{\underset{#1}{\operatorname{arg}\operatorname{max}}\;}

\usepackage[accepted]{icml2019}

\usepackage{comment}

\icmltitlerunning{Natural and Adversarial Error Detection using Invariance to Image Transformations}

\begin{document}

\twocolumn[
\icmltitle{Natural and Adversarial Error Detection\\ using Invariance to Image Transformations}



\icmlsetsymbol{equal}{*}

\begin{icmlauthorlist}
\icmlauthor{Yuval Bahat}{weiz,intern}
\icmlauthor{Michal Irani}{weiz}
\icmlauthor{Gregory Shakhnarovich}{TTIC}
\end{icmlauthorlist}

\icmlaffiliation{weiz}{Department of Applied Math \& Computer Science, Weizmann Institute of Science, Rehovot, Israel}
\icmlaffiliation{TTIC}{Toyoyta Technological Institute at Chicago, Chicago, IL, USA}
\icmlaffiliation{intern}{Part of this work was performed while the author was at TTIC}

\icmlcorrespondingauthor{Yuval Bahat}{yuval.bahat@gmail.com}
\icmlcorrespondingauthor{Gregory Shakhnarovich}{gregory@ttic.edu}
\icmlcorrespondingauthor{Michal Irani}{michal.irani@weizmann.ac.il}

\icmlkeywords{Machine Learning, ICML}

\vskip 0.3in
]



\printAffiliationsAndNotice{}  

\begin{abstract}
We propose an approach to distinguish between correct and incorrect image classifications. Our approach can detect misclassifications which either occur \emph{unintentionally} (``natural errors''), or due to \emph{intentional adversarial attacks} (``adversarial errors''),  both in a single \emph{unified framework}. Our approach is based on the observation that correctly classified images tend to exhibit robust and consistent classifications under certain image transformations (e.g., horizontal flip, small image translation, etc.). In contrast, incorrectly classified
images (whether due to adversarial errors or natural errors) tend to exhibit large variations in classification
results under such transformations. Our approach does not require any modifications or retraining of the classifier, hence can be applied to any pre-trained classifier. We further use state of the art targeted adversarial attacks to demonstrate that even when the adversary has full knowledge of our method, the adversarial distortion needed for bypassing our detector is \emph{no longer imperceptible to the human eye}. Our approach obtains state-of-the-art results compared to previous adversarial detection methods, surpassing them by a large margin.

\end{abstract}

\begin{figure}[t!] 
	\centering
        \includegraphics[width=1\linewidth]{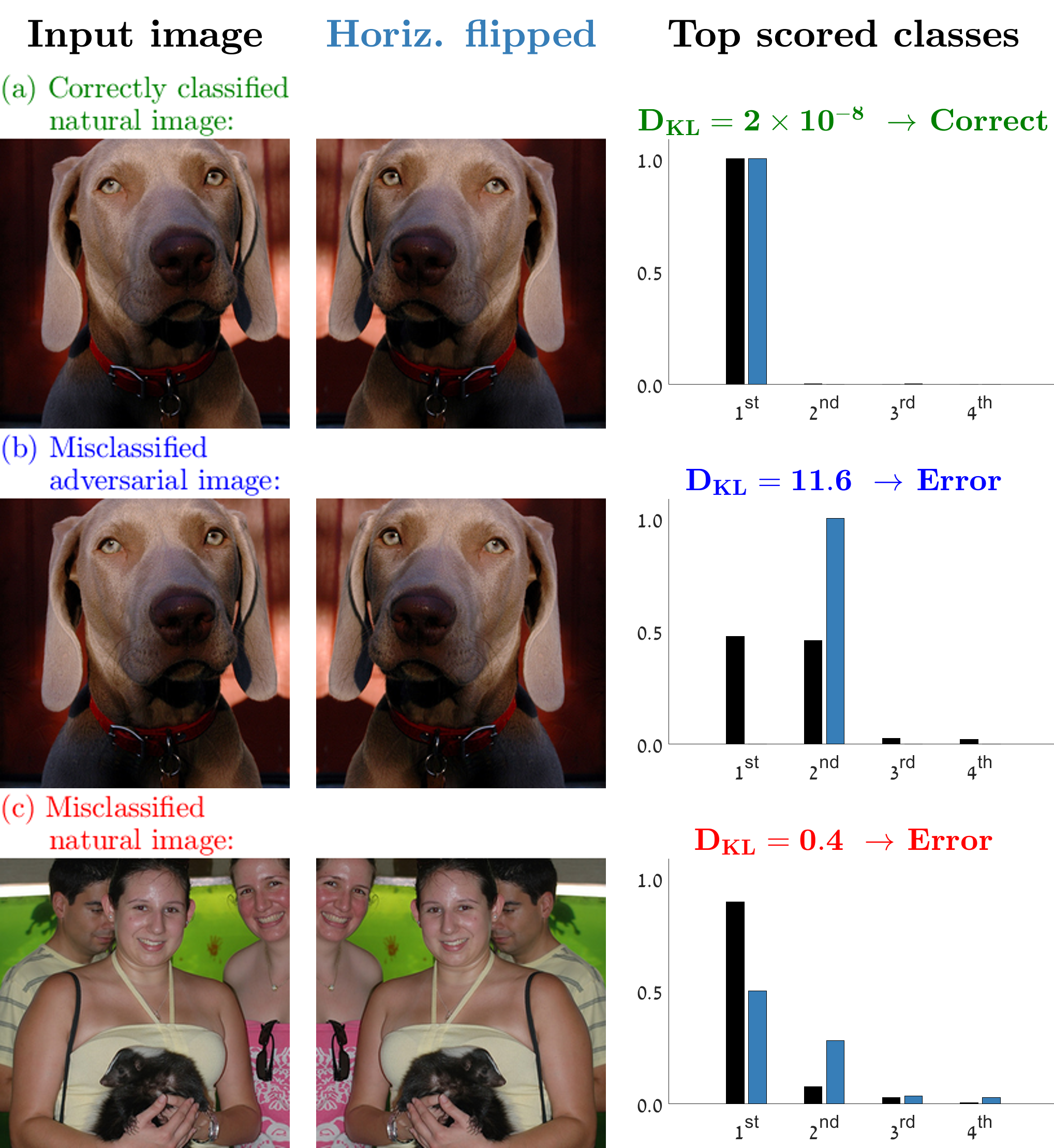}  \caption{\label{fig:transformed_images}\textbf{Invariance
            to image transformations as a proxy to classification's
            reliability. }\textit{From left to right in each row:
            input images, their horizontally flipped version and their
            (color-coded) softmax outputs corresponding to top 4 input
            image softmax values (using a pre-trained ResNet inception
            V2 classifier). Softmax values of the misclassified images
            (b and c) fluctuate across image versions, while those of
            the correctly classified image (a) are
            stable. Corresponding KL-divergence scores between softmax
            distributions (along with implied detection decision)
            indicated above each distribution allow to distinguish
            between correct and erroneous classifications.}}
	{\vspace{-0.4cm}}
\end{figure}

\section{Introduction} Despite recent progress, state of the art
recognition methods still have non-negligible error rates. For
instance, typical top-5 error rate of modern classifiers on ImageNet
is on the order of 4-5\%~\cite{ResNet,huang2017densely}. As deep
learning methods become incorporated into sensitive applications in
medical, transportation, and security domains, dealing with remaining
recognition errors becomes critical. The concern is aggravated by the
discovery of \emph{adversarial attacks} on CNN-based recognizers,
whereby errors in CNNs can be triggered on
demand~\cite{Ostrich_Szegedy2013,nguyen2015deep,madry2017PGD,CarliniW2011attack}. Such
attacks have been shown to be capable of reducing accuracy of
classifiers to arbitrarily low levels. Alarmingly, while images resulting
from these attacks manage to fool state of the art classifiers, they appear indistinguishable, to the naked eye, from ``normal'' images
classified correctly.

Traditionally, such adversarial errors and unintentional errors
(``natural errors'') have been treated as two separate problems. The
former have motivated a sequence of defense mechanisms, each in turn
defeated by subsequent modification of the attacks. The natural errors
are simply considered a fact of life, and the main avenue to deal with
them has been to improve classifier acurracy.

\noindent\textbf{Adversarial errors} Adversarial attacks are usually
categorized as \emph{black box}, when the attacker has no knowledge of
the classifier parameters (weights), and \emph{white box}, in which
the attacker has full knowledge of the classifier. We use additional
terminology to describe the level of knowledge the (white box) attacker has on the
error detector itself: \textbf{``known detector''} (KD) attackers have full
knowledge of both the classifier and the \emph{detector}, while
\textbf{``unknown detector''} (UD) attackers know only the classifier, and
do not have any knowledge of the detector. 

Additionally, one can characterize the ``strength'' of an attack by
the degree of distortion it applies to the original image in order to
``fool'' the classifier (and the detector, if relevant). The higher
the distortion the more perceptible it becomes, making it less likely
to go unnoticed by a human, thus forming a weaker (and arguably less
interesting) attack.

The rapidly expanding literature on the topic follows a ``cat and
mouse'' competition between attacks and detection/defense. Almost all
of the recently proposed detection methods have been evaluated
in~\cite{carlini2017bypassing10}, who used their strong \emph{C\&W}
attack (proposed in 2017) to attack MNIST and CIFAR-10
classifiers. They found most methods exhibit good detection
performance under the UD scenario but are easily defeated in the KD
case. They further pointed out the weakness of using (deep) learning
to construct adversarial error detectors as
in~\cite{gong2017adversarial,grosse2017statistical,metzen2017detecting}):
 ``the least effective schemes used another neural network [\ldots
 since\ldots] given that adversarial examples can fool a single classifier, it makes sense that adversarial examples can fool a classifier and detector.'' Other bypassed methods applied PCA on the images or network activations \cite{bhagoji2017dimensionality,hendrycks2016early,li2017adversarial}, or employed other statistical tests~\cite{feinman2017detecting,grosse2017statistical}.

Only the \emph{Dropout} method of~\cite{feinman2017detecting} forced the attacker to use a somewhat perceptible image distortion in order to bypass the detection, and only for the MNIST case. In the CIFAR10 case it was bypassed with an imperceptible distortion. 

\noindent\textbf{Natural error detection} The dropout method was also applied in earlier work for natural error detection~\cite{DistBasedScore_Mendelbaum2017}. An even simpler approach~\cite{Baseline_Hendrycks2016} suggests to threshold the \emph{Maximal Softmax Response} (MSR) in order to detect natural errors. 

We propose a unified framework to detect classification errors, whether natural or adversarial. 
The key idea in our approach stems from our observation that
robustness of classifiers' outputs under certain simple image
transformations (e.g. horizontal flip) is systematically different for
cases of correct classification vs. cases of misclassification. This
idea is schematically illustrated in Fig.~\ref{fig:transformed_images}. 

\begin{figure}[b!]
	\centering
\begin{subfigure}[b]{0.01\linewidth}
	\caption{\label{fig:images_types}
	}
\end{subfigure}
\begin{subfigure}[b]{0.98\linewidth}
	\centering
	\includegraphics[width=1\columnwidth]{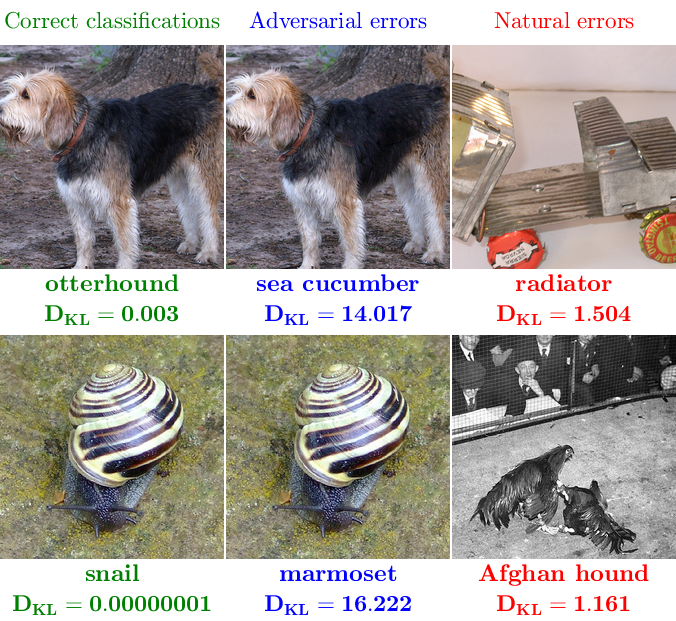}
	{\vspace{0.1cm}}
\end{subfigure}\\
\begin{subfigure}[b]{0.04\linewidth}
	\caption{\label{fig:KLD_hists}
	}
\end{subfigure}
\begin{subfigure}[b]{0.95\linewidth}
	\centering
		\includegraphics[width=1\columnwidth]{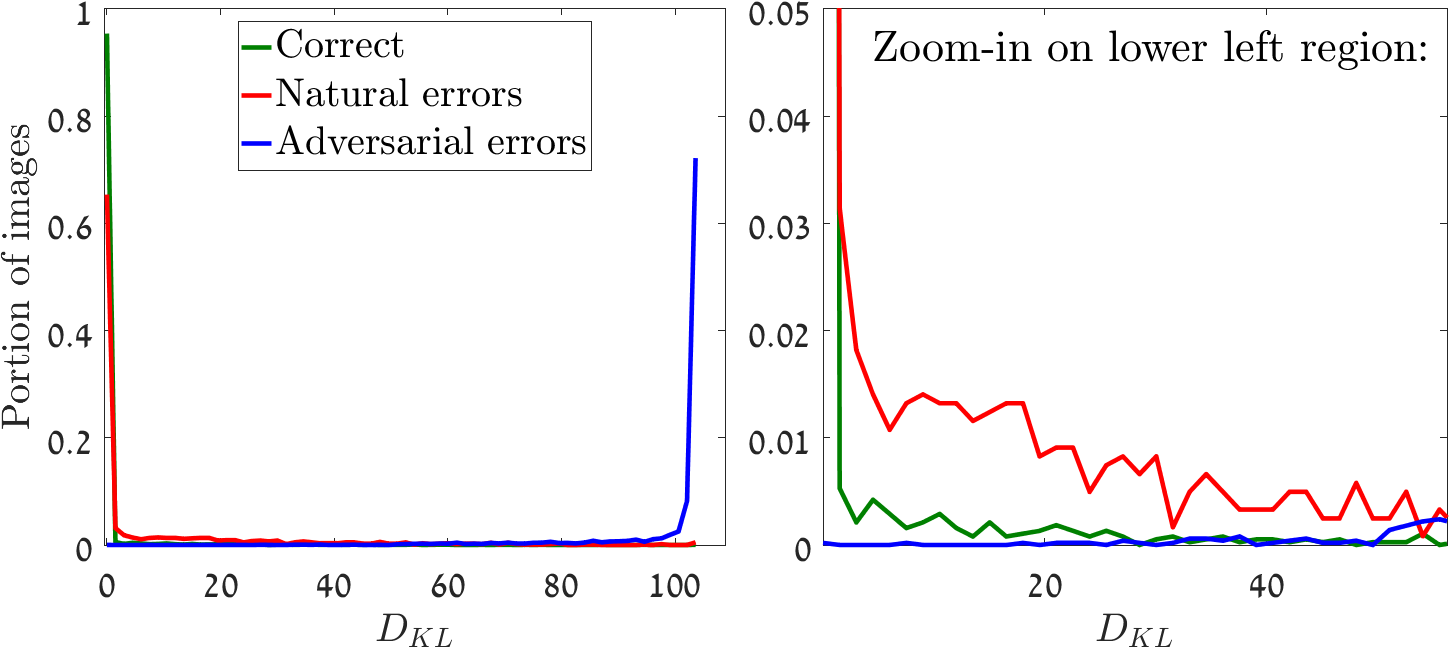}
		\end{subfigure}
	\caption{\textbf{Ability to distinguish between correctly \& incorrectly classified images.}	\label{fig:KLD_hist_and_images}\textit{(a) Examples of correctly classified (green), adversarially misclassified (blue) and naturally misclassified (red) ImageNet images. Corresponding detection scores $D_{KL}$ and predicted classes (using ResNet Inception V2) appear below each image. (b) Histograms (left) and zoom-in on lower left region (right) of $D_{KL}$ scores corresponding to 5,000 ImageNet test-set images (uniformly sampled across all classes). Separation between the different (color-coded) cases confirms the competence of the simple $D_{KL}$ score (calculated here using the horizontal flip transformation), especially for separating adversarial images (blue) from natural ones (red \& green).}}
	\vspace{-10pt}
\end{figure}
 
We measure the KL-divergence between the outputs of the classifier under image transformations.  Fig.~\ref{fig:KLD_hist_and_images} presents examples and histograms of this divergence score  $D_{KL}$ for the cases of correct, naturally incorrect and maliciously incorrect image classifications. It shows we can use this divergence score  $D_{KL}$ to detect classification errors, whether natural or adversarial (see details in Sec.~\ref{sec:invariance-concept}). This detection score can then provide a robust \emph{reject option} for classifiers, and a degree of defense against adversarial attacks.  

We show that our detector is hard for an adversary to defeat, even when the adversary has full knowledge of the detector's process. We conjecture that this is partially due to the detector's simplicity. For natural errors detection alone, we further propose to incorporate a learned \emph{Multi Layer Perceptron} (MLP) for enhanced detection performance, as we can afford to use a more sophisticated detector in that case.

Our proposed method is computationally inexpensive, and only requires
access to the inputs and outputs of the given classifier. In contrast,
the dropout method requires full control of the classifier weights at
test time, and tends to be more computationally expensive (as they run
each image 50 or 30 times through the classifier). Our method makes
use of the entire posterior distribution given by the classifier, that
was shown  to contain information valuable in the decision making
process, termed ``dark knowledge'' in~\cite{hinton2015distilling}. This is in contrast to the MSR method, that only uses the maximal posterior value.

In our evaluation, we employ some of the strongest adversarial attack methods available, like the one by \emph{Carlini \& Wagner} (2017b) (C\&W) and the \emph{Projected Gradient Descend} (PGD) attack by~\cite{madry2017PGD}. We evaluate using both \emph{targeted} attacks, where the adversary intends to change the predicted label to a specific wrong class, and \emph{untargeted} attacks, where predicting any wrong label will satisfy the adversary.

We demonstrate State Of The Art (SOTA) error detection performance in our experiments (Sec.~\ref{sec:exp}). We further show that even when the adversary has full knowledge of our method (the KD threat model), the adversarial distortion needed for bypassing our detector is no longer imperceptible to the human eye.

Our contributions can be summarized as follows:
\begin{enumerate}  \setlength\itemsep{0em}
\item We propose a \emph{unified framework} for detecting classifcation errors (whether natural or adversarial).
\item We introduce a mechanism for inferring confidence in the classification of an image from its robustness under image transformations.
\item Our detector's performance significantly improves over previous SOTA methods for detecting natural and adversarial errors (both targeted and untargeted).
\item We show that, unlike any previous method, SOTA targeted attacks,
  even with full knowledge of the detector, are forced to use \emph{perceptible distortion} in order to bypass detection.
\item Our method can be applied to any pre-trained classifier, without any modifications or retraining. We demonstrate its superior performance, even when applied to classifiers that were fine-tuned to improve performance of competing natural error detection methods.
\end{enumerate}
\vspace{-5mm}We will make our code available soon.

\section{Detecting Classification Errors} \label{sec:invariance-concept}
A common goal in building visual classifiers is that classification
output be robust under certain image transformations applied to the
image. We observe that while this is generally the case for correctly
classified images, this tends not to be true for misclassified
ones. This relationship between the correctness of a prediction and
its invariance under image transformations is the key to our approach.

We will restrict our attention to image transformations that can be
expected to occur naturally in realistic imaging conditions, without
affecting the content of the image. While we cannot provide a principled definition of the space of all such "natural" transformations, we can recognize such
transformations intuitively. For instance, horizontal (left to right) flip is natural due to lateral symmetries in the world; zooming into the image is natural since it corresponds to bringing the camera closer to the scene; etc. 

\noindent\textbf{Notation} An $N$-way classifier $F$ computes for an
input $x$ an $N$-dimensional vector of class \emph{logits}
\mbox{$\mathbf{Z}(x)\triangleq[Z_1(x),\dots,Z_N(x)]$}. Using the
softmax transformation, the logits can be converted to estimated
posterior distribution over classes $F(x)$:
\begin{equation} \label{eq:softmax}
  F_c(x)=\frac{\exp^{Z_c(x)/T}}{\sum_{c\in[N]}\exp^{Z_c(x)/T}}, \end{equation}
where $F_c(x)$ is the estimated conditional probability of $c\in[N]$ being the class of $x$. $T$ is a temperature parameter (default is $T=1$) affecting the resulting distribution's entropy, that rises with $T$ (approaching uniform distribution as $T\rightarrow\infty$, and a delta function when $T\rightarrow0$). 

A classifier prediction is made by selecting $\widehat{c}(x)=\argmax{c}F_c(x)$.
This is an error on example $x$ with label $y$ if $\widehat{c}(x)\ne y$. 

Given an image transformation $t$, we can assess the degree of
invariance of the classifier's output under this transformation as the
difference between two distributions, $F(x)$ and $F(t(x))$.
As an illustration, Fig.~\ref{fig:transformed_images} presents three
images, along with their transformed ($t$=horizontal flip)
versions. The right hand side shows the corresponding softmax outputs
for both image versions. Note that for the correctly classified image
(top row) the output is consistent across image versions, while for
the misclassified ones (middle and bottom rows) the transformation
induces significant changes in the classifier's output. It is
immaterial for this figure which class indices in the plots correspond to the
correct classes; what we are looking for is the \emph{difference}
between the $F(x)$ (black) and $F(t(x))$ (blue).

We convert this intuition into a concrete error detector by measuring
the \emph{Kullback-Leibler Divergence}\footnote{Other distance
  measures between distributions,   including Jensen-Shannon
  divergence, squared distance, or   Kolmogorov-Smirnov divergence,
  may be used; like $D_{KL}$ they exploit the ``dark knowledge''
  \cite{hinton2015distilling}, and exhibit similar behavior.}
($D_{KL}$)~\cite{kullback1959information} between the two softmax
outputs $D_{KL}\left(F(x)||F(t(x))\right)$ as a detection score. 
Figure~\ref{fig:KLD_hists} presents histograms of $D_{KL}$ scores (for $t=\textrm{hor. flip}$) corresponding to three types of classifications: correct (green), naturally incorrect (red) and adversarially incorrect (blue) classifications of ImageNet images. Adversarial images were created using the strong C\&W targeted attack on a pre-trained Inception ResNet V2 classifier~\cite{ResNet}. Note that $D_{KL}$ scores exhibit fairly good separation between adversarially (blue) and naturally (red) misclassified images, and even better separation between adversarially misclassified and correctly classified (green) images.

The
higher $D_{KL}$ is, the less stable the output of $F$ is on $x$ under
transformation $t$, and so, the less confident we are that $F(x)$ is
correct. \textbf{A binary error detector is obtained by introducing a
threshold $\tau$: if $D_{KL}\left(F(x)||F(t(x))\right)>\tau$, reject
$x$ as a misclassified image.}

In Section~\ref{sec:exp} we report on our evaluation of this simple
algorithm on the tasks of detecting targeted and untargeted
adversarial images, generated by state of the art attack methods. We found
our approach to outperform previous adversarial detection methods by a large
margin. Moreover, for the case of targeted ``known detector'' (KD)
threat model, we found that it forcest the attacker to induce a very
high and \emph{visible} image distortion in order to bypass our detector. 

We note that image transformations have also been explored as a defense mechanism by~\cite{CounterAdversarialWithManips_Guo2018}, but in a totally different way. In particular, they did not explore the divergence between classifier outputs under different transformations. In a recent work, Tian et al.~(2018) proposed to exploit the sensitivity of classification to image transformations for adversarial detection. But unlike our work their detection relies on a \emph{trained} neural network, which can be easily bypassed in an KD scenario~\cite{carlini2017bypassing10}. Moreover, they only demonstrate their method against targeted attacks on MNIST and CIFAR10 (and only for a single threshold setting).

\section{Improving Natural Error Detection} The simplicity of $D_{KL}$ score is a virtue from a security standpoint, since the lack of a complex parametric mechanism makes it harder for an adversary to circumvent it~\cite{carlini2017bypassing10} in the worst case KD scenario. This is indeed supported by our experiments (Sec.~\ref{sec:exp}) and by Fig.~\ref{fig:KLD_hists}: It indicates that the $D_{KL}$ score, based on a single transformation, suffices for detecting adversarial images. 

However, Fig.~\ref{fig:KLD_hists} also implies $D_{KL}$ is less
effective in separating \emph{natural} errors (red) from correct
classifications (green). This is partially due to the coarse binning
hiding finer separation for very low values of $D_{KL}$, but our
experiments confirm that the separation is indeed less significant
than for adversarial images. However, if we restrict our attention to
natural errors, we can afford to employ a more complex detection
mechanism to enhance detection performance (since there is no
adversary to exploit ``loopholes'' in it). Below we propose such a
mechanism that captures the rich information contained in logits
corresponding to different transformations. Here we consider a \emph{set} of image transformation $\{t_1,\dots,t_m\}$ rather than just one.

\begin{figure}[t] 
	\centering \includegraphics[width=1\columnwidth]{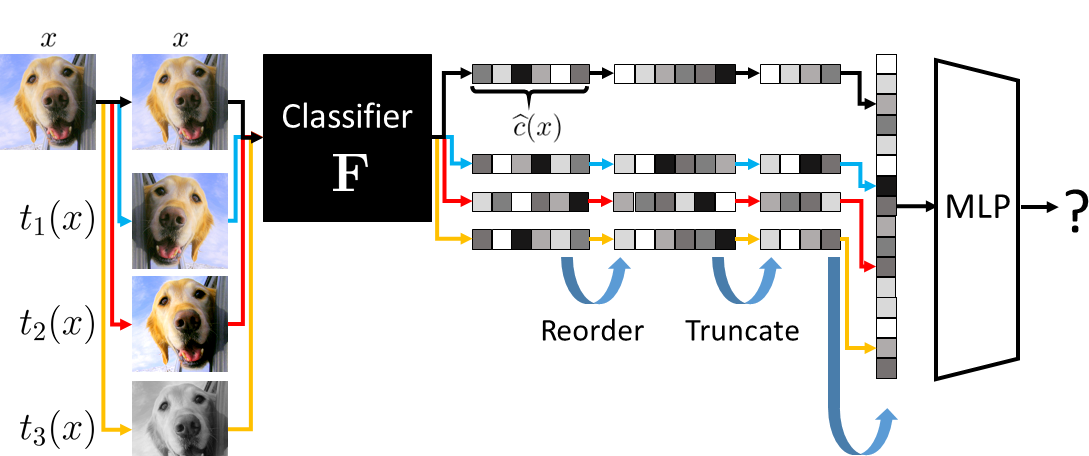}
	\caption{\label{fig:system-overview}\textbf{Overview of (optional) learned detector.} \textit{Given an image $x$ and a pre-trained classifier $F$, we feed $x$ and several (three, in this example) natural transformations of it into $F$. We jointly re-order all resulting logits vectors so that the logits vector for $x$ (the original image) is in descending order, then truncate the logits vectors to retain only the $N'$ first logits and finally concatenate them to yield the input to our detector.}}
	{\vspace{-0.2cm}} 
\end{figure}

We create a new representation for $x$ that reflects the invariance of
$F$ under these transformations, as follows (illustrated in
Figure~\ref{fig:system-overview}). Recall that $\textbf{Z}(x)$ is
the vector of logit values computed by $F$ on $x$.
\begin{enumerate}\setlength\itemsep{0em}
	\item  We jointly reorder logits vectors corresponding to the
          transformed image versions
          $\{\mathbf{Z}(t_j(x))\}_{t=1}^m$. The sorted logit order is
          the same for all transformed versions of the input image,
          and is determined by the sorting (in descending order) of the logit values of the original
          input, $\mathbf{Z}(x)$. This makes the representation independent of the predicted class $\hat{c}(x_i)$.
	\item We truncate each reordered logits vector to $N'\le N$ elements. We empirically found it reduces the detector's overfitting.
	\item We concatenate the reordered, truncated vectors into a single vector of length $(m+1)\cdot N'$.
\end{enumerate}

We then use a binary Multi Layer Perceptron (MLP) to predict
the probability of the classification $F(x)$ being incorrect. To train
it, we collect a set of category-labeled examples $(x_i,y_i)$ (ideally
from a heldout set outside of $F$'s training data), and for each $x_i$
compute the logits vectors corresponding to the original input
$\mathbf{Z}(x_i)$ and its transformed versions
$\{\mathbf{Z}(t_j(x_i))\}_{t=1}^m$. These logits constitute the MLP's
input. We then assign this input with its corresponding \emph{error
label} $e_i$, set to $1$ if $y_i\neq F(x_i)$ and -1 otherwise. See
Section~\ref{sec:exp} for details on architecture and training of the MLP.



Note that in a scenario involving both natural and adversarial images,
we could theoretically start by rejecting adversarial images using our
simple $D_{KL}$-based mechanism. Then, assuming our detector is no
longer susceptible to adversarial attacks, we could proceed to use
this logits-based MLP for enhanced natural error detection. However,
this mode of operation may introduce additional vulnerabilities, for
instance adversary fooling the MLP detector into rejecting correctly
classified images as erroneous.

\noindent\textbf{Relation to data augmentation}
Image transformations are commonly employed for augmenting the
training set of a classifier.
This may raise a question: how can we
benefit from invariance to a transformation if the classifier,
ostensibly, is trained to be invariant to it? A key observation here
is that data augmentation does not necessarily force the classifier to
learn \emph{invariance} to transformations of a given image - the classifier may simply
learn that the different transformed versions correspond to different
\emph{instantiations} of the same class,
encouraging each one \emph{separately} to output the correct class. In
contrast, our method infers confidence in a prediction from actual
invariance of classifiers' 
outputs on transformed versions of an image. In fact, results in Fig.~\ref{fig:KLD_hists} were obtained using the horizontal flip transformation, for classifiers that included horizontal flip in their data augmentation procedure during training.
Moreover, comparing our method's performance on two CIFAR10
classifiers trained with vs. without horizontal flip augmentation
showed no meaningful difference in our ability to reject errors. This suggests our method's performance is independent of the classifier's training data augmentation procedure.


\section{Experiments} \label{sec:exp} We evaluate our error detection method using four data-sets: CIFAR-10~\cite{krizhevsky2009CIFAR}, STL-10~\cite{STL10}, CIFAR-100~\cite{krizhevsky2009CIFAR}, and ImageNet~\cite{ImageNet}, which vary in terms of both number of examples and the number of classes. This is to confirm that the proposed method generalizes across task sizes. We demonstrate results with different classifiers, including CIFAR10 classifiers from \cite{CarliniW2011attack} and \cite{carlini2017bypassing10}, classifiers trained by \cite{DistBasedScore_Mendelbaum2017} and the competitive Inception-ResNet-V2 classifier \cite{ResNet}.

We employ the \emph{Receiver Operating Characteristic} (ROC) curve and the corresponding \emph{Area Under ROC Curve} (AUROC) to allow thorough evaluation of various detection policies (e.g. low mis-detection, low false-detection).

\begin{figure}[!tbh]
	\centering
	\includegraphics[width=1.1\columnwidth]{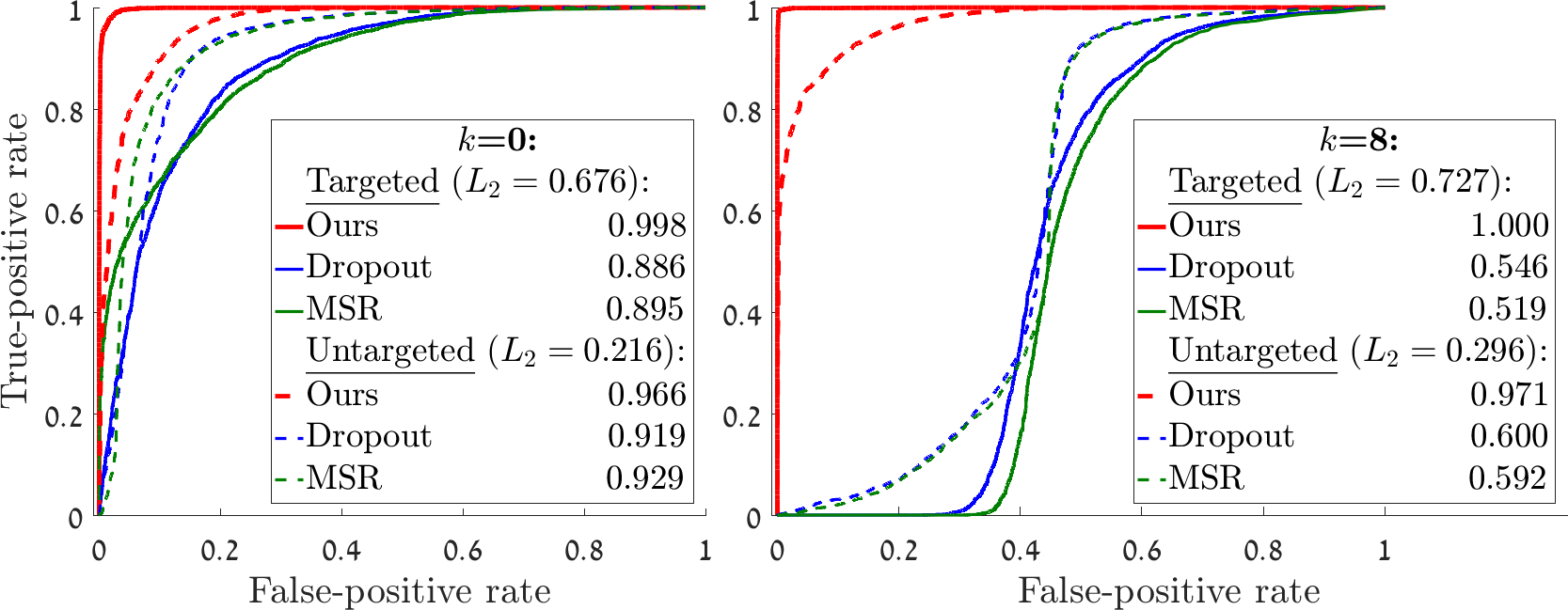}
	\vspace{-10pt}
	\caption{\label{fig:UD_ROC}\textbf{``Unknown detector'' (UD)
            threat model.} \textit{ROC curves corresponding to
            detecting targeted (solid line) and untargeted (dashed
            line) C\&W attacks on ImageNet \cite{ImageNet} images
            (using the same classifier as in
            Fig.~\ref{fig:KLD_hists}). The L2 norms in the legend
            indicate the average (per pixel) norm of the perturbation
            produced by the attacker; pixels are in [0,255] range.
Values in legend correspond to the AUROC (Area Under ROC) of the
different methods. We use our method (red) with $t=$hor. flip and
compare it to the dropout (blue) and MSR (green) methods. Increasing
the attack's confidence parameter from $k=0$ (left) to $k=8$ (right)
hardly affects the adversarial noise norm ($L_2$ in the legends), but significantly impairs performance of both competing
methods. In contrast, our method still achieves AUROC$\approx$1,
indicating nearly perfect detection.}}
\end{figure}
 
\subsection{Adversarial image detection} \label{sec:exp-adversarial}
We follow the framework put forth by~\cite{carlini2017bypassing10}
which was able to bypass ten adversarial detection methods, and
evaluate our detector using their attack~\cite{CarliniW2011attack}
(C\&W)\footnote{We further experimented with the PGD
  attack~\cite{madry2017PGD}. but omit these experimental details, as we found PGD to be a much weaker attack (compared to C\&W), easily detected or prevented by our method in both KD and UD scenarios.  We hence focus on our experiments with the more challenging C\&W attack.} under the two
different threat models, ``Unknown Detector'' (UD) and ``Known
Detector'' (KD).
We compare our method to the simple \emph{Maximal Softmax Response}
(MSR) method \cite{Baseline_Hendrycks2016} and to the dropout method
\cite{feinman2017detecting}, that was found to be the most effective
by~\cite{carlini2017bypassing10}. The methods are compared on 10,000
CIFAR-10 validation set images and on 5,000 images sampled from ImageNet
validation set.

\subsubsection{``Unknown Detector'' (UD) Performance}
Figure~\ref{fig:UD_ROC}
shows ROC curves and their corresponding Area Under ROC values (``AUROC'') on the task of detecting adversarial images. The reported results are on ImageNet, using a horizontal flip transformation.

We run both
targeted and untargeted
attacks (in the targeted case we randomly assign target labels).
We find our method ranks best in all comparisons, nearing perfect detection capabilities (AUROC$\approx1$).

\begin{figure}[!h]
	\centering
	\includegraphics[width=1\columnwidth]{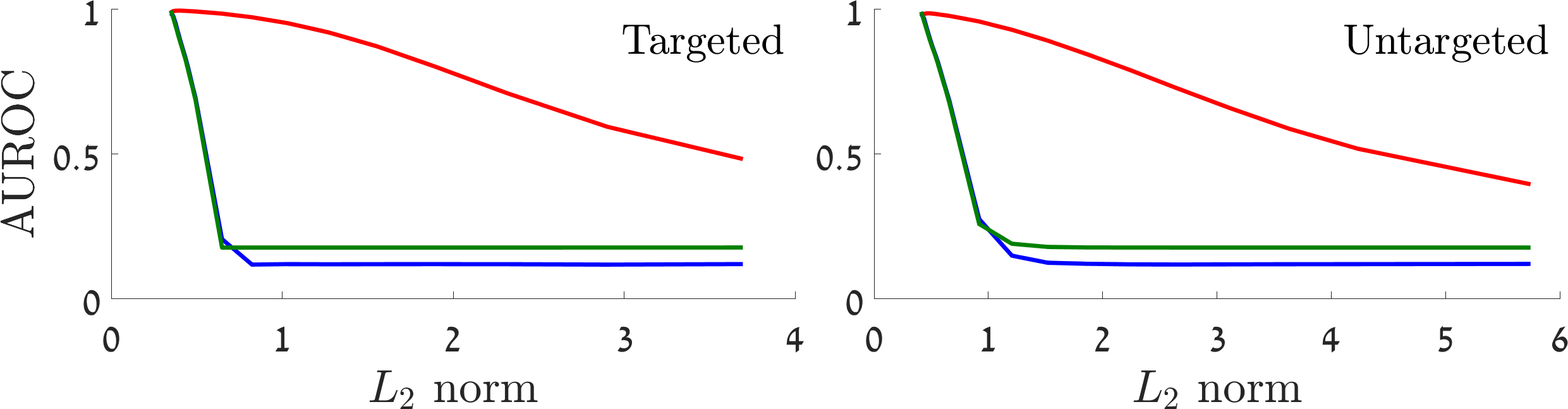}
	\vspace{-10pt}
	\caption{\label{fig:AUROC_vs_CarliniConf}\textbf{Detection performance vs. adversarial distortion.} \textit{Increasing the C\&W attack's confidence parameter $k$ results in more confidently misclassified images with increased transferability, while increasing image distortion. Unlike for our method (red), detection performance (AUROC) on CIFAR10 images drop drastically for the dropout \cite{feinman2017detecting} (blue) and MSR \cite{Baseline_Hendrycks2016} (green) methods as $k$ (and adversarial distortion) increase, in both targeted (left) and untargeted (right) UD attacks.}}
\end{figure}

The  C\&W attack employed in these experiment has a tunable parameter
``confidence'' $k$ that determines how confidently should classifier
$F$ misclassify the resulting adversarial image. Higher values
lead to more confidently misclassified examples with better
transferability across classifiers
~\cite{CarliniW2011attack}, at the cost of increased image
distortion (quantified here using the $L_2$ norm). The experiments in
Fig.~\ref{fig:UD_ROC} were conducted using either the minimum attack
confidence value $k=0$ (left) or a slightly higher one, $k=8$
(right). Note that despite the imperceptible increase in image
distortion, the advantage of our method becomes even clearer in the
higher confidence case (right). We see the same phenomenon on CIFAR10
(using the classifier from
\cite{carlini2017bypassing10}). Fig.~\ref{fig:AUROC_vs_CarliniConf}
depicts the resulting AUROC vs. average image distortion norm
(controlled indirectly by varying confidence $k$ of the attacker) for the targeted (left) and untargeted (right) attacks. It shows our method can detect highly confident, transferable, adversarial images well after the competing methods drop to chance level accuracy (AUROC around or below $0.5$). 

\begin{table}[h]
	\caption{\label{tab:different_transformations}\textbf{Detecting UD attacks using different transformations.} \textit{AUROC values reflect detection performance for {\bf untargeted} UD C\&W attacks (with $k=0$) on the ResNet Inception V2 ImageNet classifier and the CIFAR-10 classifiers from~\cite{carlini2017bypassing10} (``ResNet32'') and~\cite{papernot2016distillation} (``basic''). 
	}}
      \vspace{5mm}\begin{tabular}{|c||c|c|c|}
	\hline
		&\shortstack{CIFAR-10\\(basic)}&\shortstack{CIFAR-10\\ (ResNet32)}&\shortstack{ImageNet\\(top~5)}\\\cline{2-3}
		\hline
		Hor. flip&0.936&0.976&0.966\\
		Gamma$^{0.6}$&0.906&0.954&0.941\\
		Zoom$\times1.05$&0.932&0.974&0.964\\
		\hline
	\end{tabular}
\end{table}
 Finally, in
Tab.~\ref{tab:different_transformations} we examine the effect of
transformation choice $t$ on performance. We tested the ability to
detect untargeted C\&W attacks using the horizontal flip,
Gamma correction
and zoom-in transformations, and
found they all yield excellent detection performance, often close to
perfect (AUROC close to $1$). We further experimented with averaging
or taking the maximal $D_{KL}$ score over several transformations, and
found it yields similar performance to that using individual transformations.

\subsubsection{``Known Detector'' (KD)
  performance} \label{sec:exp-full_knowledge} We next present
experiments with the KD threat model, where an adversary attempts to
simultaneously deceive the classifier and bypass the detection
mechanism. It should be noted that any detection method can eventually
be bypassed, since the attacker could simply replace an image with an
alternative natural image from another class, thus achieving desired
``wrong'' classification. However, such trivial ``attacks'' are easily
visible to a human observer, whereas imperceptible attackes are both
more powerful and more interesting from the perspective of
understanding vulnerabilities of current visual recognition.

We show that to overcome our detector even with moderate level of success, the attacker needs to modify the original image to a degree that is \emph{perceptible to the human eye}, in contrast to the much lower distortion imposed by previously proposed detection methods. 
To this end, we employ the commonly used $L_2$ and $L_\infty$ norms
(measuring the squared change averaged over pixels, and maximal change
over pixels,
respectively) to quantify this distortion, and conduct our experiments using a targeted C\&W attack \cite{CarliniW2011attack} that explicitly minimizes this $L_2$ norm while manipulating images.

The AUROC curves in Fig.~\ref{fig:AUROC_vs_CarliniConf} show that attacking with higher confidence parameter $k>0$ is enough to deceive both the dropout \cite{feinman2017detecting} and MSR \cite{Baseline_Hendrycks2016} methods, while imposing imperceptible distortion. However, this approach is clearly ineffective against our method, that requires much higher, perceptible, distortion in order to be breached. 

To make the attack more credible, we adopted the following successful approach from \cite{carlini2017bypassing10} as our KD attack: We define a new function $G(x)$ that represents the combination of a given $N$ classes classifier $F$ with logits $\{Z(x)_i\}_{i=1}^N$, and our detectors' output score, $D_{KL}(x)$. 
$G$ is a classifier with $N+1$ classes. It classifies image $x$ according to $F$ (i.e. as belonging to one of the $N$ classes) whenever our detection score is below a pre-defined detection threshold $\tau$, and classifies $x$ as an ``adversarial'' (class $N+1$) otherwise. 
We can then simulate the KD threat model by running a targeted C\&W
attack on $G$ instead of $F$, producing images that manage to fool
classifier $F$, while keeping clear of the error detector threshold
$D_{KL}(x)<\tau$. We use $\tau$ corresponding to $1\%$ false detection rate (FPR) throughout our experiments.
\begin{table}[h]
	\begin{center}
	\caption{\label{tab:WB_configurations}\textbf{``Known
            detector'' (KD) threat model, targeted attacks (ImageNet).}  \textit{From left to right: Detector configuration, bypassing rate corresponding to $1\%$ FPR and average imposed $L_2$ and $L_\infty$ norms (using a 0-255 scale). Comparing the UD attack (row 0) with KD attacks using several transformations with temperature $T=1$ (rows 1-3) vs. $T\approx 0.15$ (rows 4-6). Distortion norms computed over 250 ImageNet images. We denote zoom-in times $1.03$ by `zoom' and image translation of $x,y$ (in the corresponding axes) by `shift$_{x,y}$'.}}
		\begin{tabular}{|l|c|c|c|}
			\hline
			\multicolumn{1}{|c|}{Configuration}&\shortstack{Bypass\\@1\%FPR}&$||\cdot||_2$&$||\cdot||_\infty$\\
			\hline\hline
			\rowcolor{lightgray}(\rownumber)~UD:~$t=$~hor. flip&2\%&0.75&12.39\\
			\hline\hline
			(\rownumber)~$t=$~zoom&97\%&1.71&20.15\\
			\hline
			(\rownumber)~$t=$~hor. flip&100\%&1.64&18.48\\
			\hline
			(\rownumber)~$t=$~shift$_{\frac{1}{2},\frac{1}{2}}$&93\%&1.2&16.32\\
			\hline\hline
			(\rownumber)~$t=$~zoom$,T=0.16$&55\%&6.84&47.5\\
			\hline
			(\rownumber)~$t=$~hor. flip$,T=0.12$&53\%&\textbf{7.54}&\textbf{51.45}\\
			\hline		
			(\rownumber)~$t=$~shift$_{\frac{1}{2},\frac{1}{2}},T=0.15$&84\%&4.17&31.79\\
			\hline
	\end{tabular}
	\end{center}
\end{table}

\begin{figure}[t!]
	\centering
        \includegraphics[width=1\columnwidth]{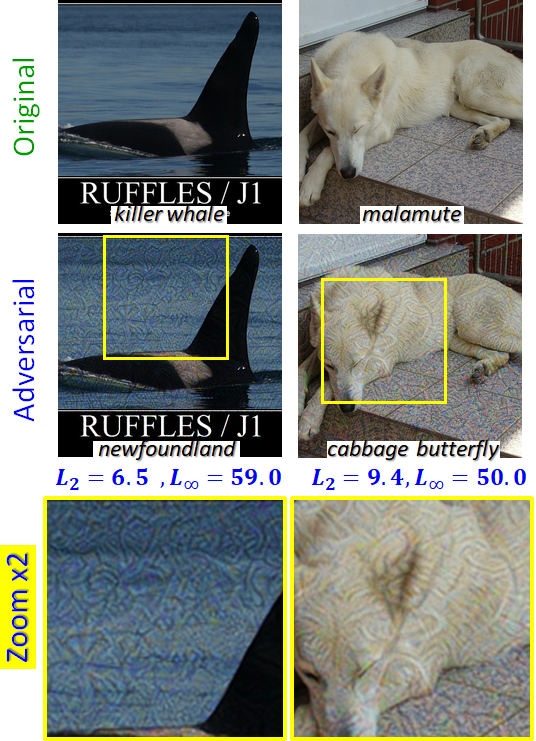}
	\caption{\label{fig:WB_images}\textbf{Our detector forces
            large, perceptible perturbations in adversarial images.}
          {\it ``Known detector'' attack; Original images (top) and
            their successfully deceiving manipulated versions
            (middle). Bypassing our detector (with $t=$ hor. flip,
            $T=0.12$) under the challenging KD threat model requires
            significant image distortion, that is easily noticeable by
            the naked eye -- see slightly zoomed in regions for
            emphasis (bottom). Class prediction and distortion norms reported below each image.}}
	{\vspace{-0.2cm}}
\end{figure}

We find the temperature $T$ in Eq.~\eqref{eq:softmax} to have a dramatic
effect on the detection performance. Setting $T=1$ (rows 1-3 in
Tab.~\ref{tab:WB_configurations}) allows the attack to
evade detection in almost $100\%$ of images while imposing only
slight, imperceptible, increase in distortion. However, lowering the
softmax temperature parameter $T$ (Eq.~\ref{eq:softmax}) dramatically
increased our method's robustness. Rows 4-6 corresponding to $T\approx
0.15$ show that our method prevents almost half the adversarial attacks, and requires an
order-of-magnitude increase in average distortion (row 5) of those
images that do bypass it successfully, compared to the UD attack (row
0). This degree of distortion is no longer imperceptible to the naked
eye, as exemplified in Fig.~\ref{fig:WB_images}. We emphasize again
that in these KD experiments, the attacker has access to full
information about the detector, including the value of $T$.

To understand the effect of the
temperature parameter $T$ in Eq.~\ref{eq:softmax}, recall that in the
KD scenario the adversary attacks the combined classifier-detector
model $G$. This means the adversary creates images that simultaneously
manage to deceive the classifier, $F(x)\neq y$, and maintain small
distance (measured using $D_{KL}$) between the outputs corresponding
to the original and transformed images, $F(x)$ and $F(t(x))$,
respectively. Setting lower $T$ focuses this distance computation on
the few top ranking classes, rather than evenly weighing the entire
softmax distribution. This makes the adversary's task considerably
more difficult.

We have also experimented with \emph{untargeted} KD attacks (an easier task for
the attacker, with much more flexibility). Here too we observed the
detector forcing higher perturbations in the images to bypass the detector, but not rising to
the level easily perceptible to humans ($L_2=1.4$, $L_{\infty}=19.5$). However ImageNet
has many closely related classes, e.g., similar breeds of dogs. This
high inter-class similarity renders
untargeted attacks in this domain somewhat less interesting.


\subsection{Natural error detection}
We next evaluate the enhanced natural error detection performance. Our
detector is a fully connected network with two hidden layers of width 30, each followed by RELU nonlinearity, and a batch normalization layer. We train it using asymmetric cross-entropy loss (reweighting examples to correct for the imbalance in number of correct vs. error training examples) and dropout probability $0.5$ in both hidden layers.

For training it, we use images from the original validation-set, by
splitting it into training and validation subsets for our
detector.\footnote{We avoid training our MLP on images that were used
  for the classifiers' training, to avoid train/test disparity since
  the classifier had a chance to optimize its output on the training
  images. Instead we train the detector on a subset of the validation set. The subset assignment for each data-set is consistent across all our experiments, and will be made public, along with our code.} When performing the pre-processing (described in Sec.~\ref{sec:invariance-concept}) for the STL-10, CIFAR-100 and ImageNet classifiers, we truncate each input logits vector and leave only the top $N'=5/10/20$ ranked classes of the input image, respectively.

We augment our detector's training set by randomly applying horizontal flip and random brightness and contrast adjustments (and for ImageNet also random cropping). This data augmentation pipeline is performed \emph{before} applying our (fixed) transformations $t$.

\begin{table}[t]
	\begin{center}
	\caption{\label{tab:error_datect_methods_Mendelbaum_AUROC_clsfr}\textbf{Detecting natural errors.} \textit{Comparing detection performance (AUROC) of our method with the MSR, dropout and SOTA DBC methods. We detect using either hor. flip based $D_{KL}$ score or an MLP (based on the hor. flip, Gamma correction, contrast modification, gray-scale conversion and small horizontal blur transformations). Performance is compared using classifiers pre-trained by the DBC work, using either solely cross entropy loss (Regular) or its combination with an adversarial loss term (AT), for their improved performance.}}
			\vspace{2mm}\begin{tabular}{|c||c|c|c|c|}
			\hline
			\multirow{2}{*}{}& \multicolumn{2}{c}{STL-10} & \multicolumn{2}{|c|}{CIFAR-100}\\\cline{2-5}
			& Regular & AT & Regular & AT\\
			\hline
			MSR
			&0.806&0.813&0.834&0.842\\
			Dropout
			&0.803&0.809&0.834&0.847\\
			DBC 
			&0.786&0.866&0.782&0.858\\
                          \hline
                          $D_{KL}$ (Ours)&0.814&0.801&0.835&0.825\\
			MLP (Ours)&\textbf{0.846}&\textbf{0.868}&\textbf{0.864}&\textbf{0.869}\\
			\hline
		\end{tabular}

	\end{center}\vspace{-1em}
\end{table}
We evaluate our method on the STL-10 and CIFAR-100 datasets, and compare it to MSR \cite{Baseline_Hendrycks2016}, dropout \cite{feinman2017detecting} and the SOTA \emph{Distance Based Confidence} (DBC) method by Mandelbaum and Weinshall (2017). DBC detects natural errors by accessing the classifier's internal activations. 

Table~\ref{tab:error_datect_methods_Mendelbaum_AUROC_clsfr} present Area Under ROC (AUROC) values obtained using two types of pre-trained classifiers of~\cite{DistBasedScore_Mendelbaum2017}. These two types differ in the loss function utilized for their training: The \emph{Regular} classifier used only the cross entropy loss. The \emph{AT} classifier was fine-tuned using the adversarial loss term of \cite{Adversarial_Goodfellow2014}\footnote{Another comparison on a third loss term proposed in~\cite{DistBasedScore_Mendelbaum2017} yielded similar results, omitted here for lack of space. When using their classifier, we follow~\cite{DistBasedScore_Mendelbaum2017} and pre-process the original and transformed images by global contrast minimization and ZCA whitening.}, that was found to improve the performance of DBC.

Using our basic $D_{KL}$ score yields favorable performance on the \emph{regular} classifiers. However, using our MLP detector achieves the best performance on all classifiers and datasets. Note that while our detector can be used on any given, pre-trained, classifier, it achieves SOTA performance even on classifiers that were modified by the DBC method in its favor (the AT configuration).

\noindent\textbf{ImageNet} Finally, we evaluated our natural error detection performance (with and without using an MLP) when applied to the ILSVRC-2012 ImageNet classification task. We used a pre-trained Inception-ResNet-v2 model \cite{ResNet} that achieves $81\%$ and $95.5\%$ top-1 and top-5 accuracies, respectively. As before, we used part of ImageNet's validation set to train our detector ($20\%$ in this experiment), and evaluated its performance on the remaining part. 
Table~\ref{tab:error_detect_ImageNet} compares AUROC values corresponding to ours and the MSR~\cite{Baseline_Hendrycks2016} methods. Due to the scale of ImageNet, we did not compare to methods that require either extensive re-training (DBC) or extensive computation (dropout).

\begin{table}[!thb]
	\begin{center}
	\caption{\label{tab:error_detect_ImageNet}\textbf{Detecting natural errors on ImageNet.} \textit{Comparing detection performance (AUROC) of our method using either $D_{KL}$ score or an MLP (both based on the same transformations as in Tab.~\ref{tab:error_datect_methods_Mendelbaum_AUROC_clsfr}), with the MSR method. Tested on a pre-trained Inception-ResNet-V2 classifier.}}
        \vspace{2mm}\begin{tabular}{|c||c|c|}
			\hline
			& Top-1 & Top-5\\
			\hline
			MSR&0.842&0.806\\
                  \hline
                  $D_{KL}$ (Ours)&0.852&0.823\\
			MLP (Ours)&\textbf{0.875}&\textbf{0.884}\\
			\hline
		\end{tabular}
	\end{center}\vspace{-2mm}
\end{table}

\section{Conclusions}\label{sec:conclusion} We devise a simple \&
robust classification error detector based on our observation that
incorrect predictions correspond to less stable classifier outputs
under a set of image transformations. We demonstrate the effectiveness
of this approach for rejecting adversarial examples, under a variety
of attack scenarios, including the most challenging Known Detector
(KD) attack by the Carlini \& Wagner method. We further propose to enhance detection of natural errors by training an invariance based MLP. Beyond the immediate applications to increase robustness and reduce classifiers' vulnerability, the success of our detection methods suggests further study of invariance of convolutional networks under image transformations, and the potential role this invariance may play in improving recognition.

\section*{Acknowledgments}
This material is based in part on research sponsored by the US Air Force Research Laboratory and DARPA under agreements number FA8750-18-2-0126 and FA9550-18-1-0166. The U.S. Government is authorized to reproduce and distribute reprints for Governmental purposes notwithstanding any copyright notation thereon. 
This project has received funding from the European Research Council (ERC) under the European Union's Horizon 2020 research and innovation programme (grant agreement No 788535).
\bibliography{natural_and_adversarial_error_detection}
\bibliographystyle{icml2019}


\end{document}